\pdfoutput=1

\documentclass[11pt]{article}

\usepackage{acl}

\usepackage{times}
\usepackage{latexsym}
\usepackage{graphicx}
\usepackage{enumitem}

\usepackage{CJKutf8}

\usepackage[T1]{fontenc}

\usepackage[utf8]{inputenc}

\usepackage{microtype}

\usepackage{inconsolata}

\usepackage{float} 
\usepackage{tabularx}
\usepackage{multirow}
\usepackage{arydshln}

\newcommand{\DATASET}{ModePoem}
\newcommand{\GP}[1]{$\mathcal{P}_{#1}$}
\newcommand{\HP}[1]{$\mathcal{H}_{#1}$}

\newcommand{\PARASPACE}[1]{\setlength{\baselineskip}{0.7\baselineskip}#1}
\newcommand{\MYFONTSIZE}[1]{\PARASPACE{{\fontsize{8}{9}\selectfont {{#1}}}}}

\title{What is the Best Way for ChatGPT to Translate Poetry?}

\author{
       Shanshan Wang$^1$~~~~
        Derek F. Wong$^1\thanks{Corresponding Author}$~~~~
        Jingming Yao$^2$~~~~
        \textbf{Lidia S. Chao$^1$}~~~~
       \\
    $^1$NLP$^2$CT Lab, Department of Computer and Information Science, University of Macau \\
    {nlp2ct.shanshan@gmail.com, \{derekfw,lidiasc\}@um.edu.mo} \\
     $^2$Department of Portuguese, Faculty of Arts and Humanities, University of Macau\\
       {jmyao@um.edu.mo} \\
    }

\begin{document}
\maketitle
\begin{abstract}
Machine translation (MT) has historically faced significant challenges when applied to literary works, particularly in the domain of poetry translation. The advent of Large Language Models such as ChatGPT holds potential for innovation in this field. This study examines ChatGPT's capabilities in English-Chinese poetry translation tasks, utilizing targeted prompts and small sample scenarios to ascertain optimal performance. Despite promising outcomes, our analysis reveals persistent issues in the translations generated by ChatGPT that warrant attention. To address these shortcomings, we propose an Explanation-Assisted Poetry Machine Translation (EAPMT) method, which leverages monolingual poetry explanation as a guiding information for the translation process. Furthermore, we refine existing evaluation criteria to better suit the nuances of modern poetry translation. We engaged a panel of professional poets for assessments, complemented evaluations by using GPT-4. The results from both human and machine evaluations demonstrate that our EAPMT method outperforms traditional translation methods of ChatGPT and the existing online systems. This paper validates the efficacy of our method and contributes a novel perspective to machine-assisted literary translation.\footnote{https://github.com/NLP2CT/Poetry-Translation} 

\end{abstract}

\begin{figure*}[t]
    \centering
    \includegraphics[width=1.0\textwidth]{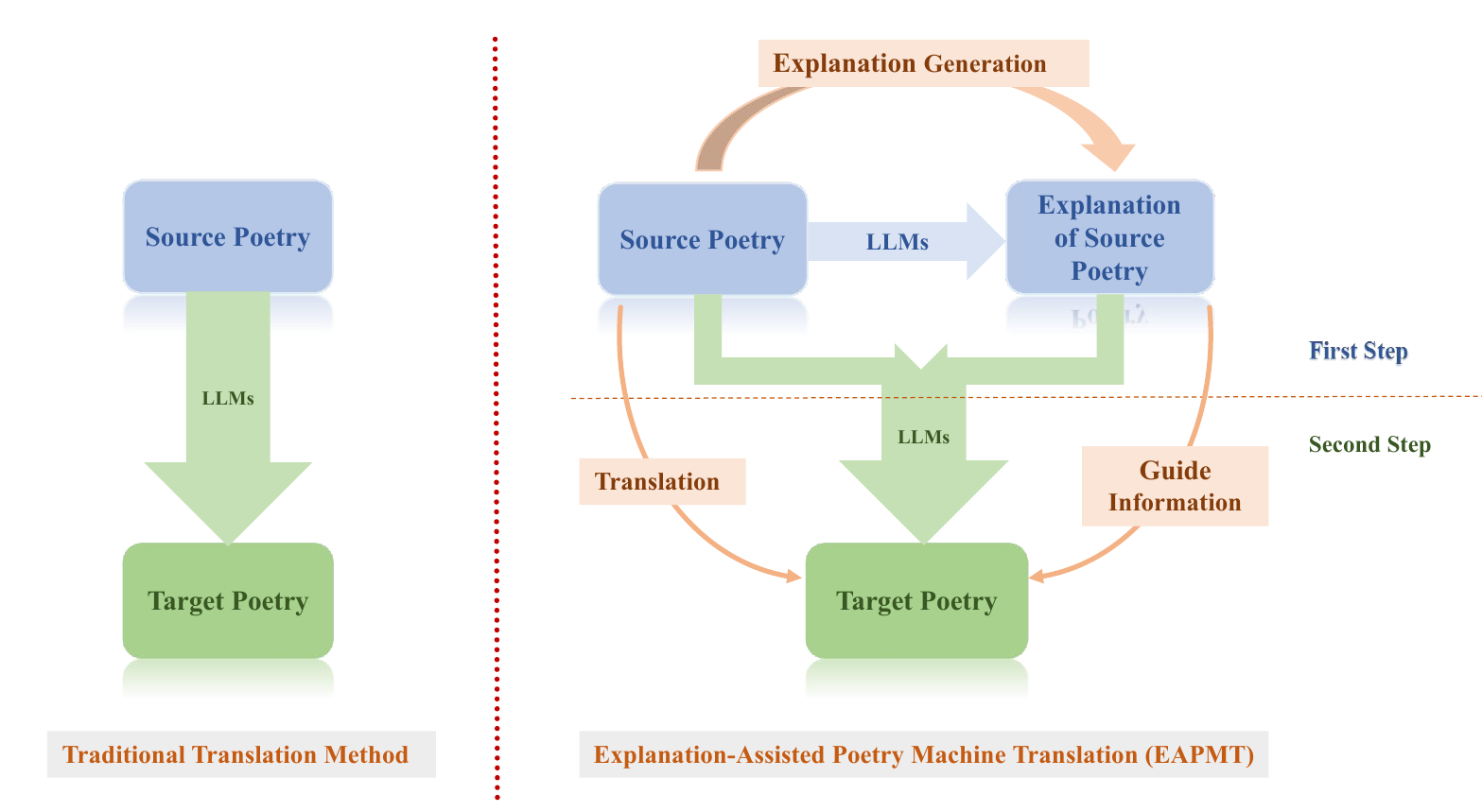}
    \caption{Comparison between the framework of the traditional translation method and the proposed Explanation-Assisted Poetry Machine Translation (EAPMT).}
    \label{EAPMT}
\end{figure*}



\section{Introduction}
\textit{Foreign poems translated into Chinese still have to be like poems.}
\begin{flushright}—— \citet{GuoMoruo}\end{flushright}

Poetry translation is widely regarded as one of the most challenging tasks in the field of translation. When translating a foreign poem into Chinese, the resulting text should still be recognizable as a poem~\cite{GuoMoruo, bianzhilin1989translation}. In discussing poetry, we often refer to specific genres or styles \cite{stevenson1957poem}. Unlike classical poetry, the term ``modern'' in modern poetry refers to the poetic styles of the 20th and 21st centuries, which represent a significant departure from traditional forms.
The primary characteristic of modern poetry is its embrace of freedom and lack of restrictions \cite{awan2015new}. Modern poetry is specifically characterized by open forms, diverse genres, a break from conventional narratives, and innovative language combinations \cite{GuoMoruo, wang2006guoview, skerratt2013form}. Unlike classical poetry, rhythm is no longer an essential feature of modern poetry. Consequently, when translating modern poetry, it is not necessary to adhere to the original poem's rhythm \cite{duanjinyan2008research}. However, the poeticity must not be overlooked; the poetic essence of the source poem must be preserved throughout the translation process \cite{GuoMoruo, skerratt2013form}.

For Chinese poetry, there are significant differences between various genres. Classical Chinese poetry is characterized by strict constraints on format, meter, sentence length, and rhyme. In contrast, modern poetry is free from these constraints and breaks away from the rigid structures of classical poetry \cite{skerratt2013form, awan2015new}.



Previous work has successfully applied machine translation to poetry, but these poems typically have clear format or rhyme restrictions \cite{genzel2010poetic,  ghazvininejad2018neural, chakrabarty2021don, chakrawarti2022machine, song2023towards}, which differ significantly from modern Chinese poetry. Recent research has taken an innovative approach by first obtaining an initial translation of the input prose using traditional Neural Machine Translation (NMT) methods. This initial translation is then mapped to a set of masked sequences via a designed heuristic method. Finally, these sequences are used to generate poetry translations through a pre-trained Masked Language Modeling (MLM) technique \cite{khanmohammadi2023prose2poem}.

Another study compared the differences between machine translation and human translation of Arabic poetry into English \cite{alowedi2023artificial}. The authors concluded that machine translation is not suitable for translating Arabic poetry into English as it fails to comprehend the socio-cultural background of poetry creation and the contextual nuances, particularly the genre-specific elements.

Promisingly, the artificial intelligence chatbot ChatGPT, released by OpenAI, has demonstrated excellent performance across various tasks and domains, including translation tasks \cite{jiao2023chatgpt, zhan2024prefix, pang2024anchor, lan2024focus}.
Although previous work has studied the performance of ChatGPT on translation tasks \cite{hendy2023good, jiao2023chatgpt, yang2023human}, and recent studies have explored the application of ChatGPT to poetry-related tasks, these investigations primarily focused on poetry generation. For instance, recent research examined the effectiveness of ChatGPT-4 in generating Arabic poetry and found the results to be unsatisfactory \cite{antar2023effectiveness}. The study highlighted several issues with the text generated by ChatGPT-4, including poor language quality, superficial content, lack of emotion, inconsistent speech, inappropriate word choices, and an ease of recognition by human evaluators.

Unlike previous work, this paper focuses on the capabilities of ChatGPT in translating English poetry into modern Chinese poetry. We explored optimal strategies for utilizing ChatGPT to translate poetry and evaluated its maximum performance in this specific task.
Inspired by \citet{jiao2023chatgpt} and \citet{gao2023design}, we investigated ChatGPT's performance on modern poetry translation tasks by designing appropriate prompts and providing example shots for the model.

Experimental results demonstrate the effectiveness of our designed prompts. Despite these promising outcomes, our analysis reveals persistent issues in the translations generated by ChatGPT that warrant further attention. Consequently, we propose a new poetry translation method called Explanation-Assisted Poetry Machine Translation (EAPMT). Our method leverages the explanation of monolingual poetry as guidance information to achieve high-quality translations from English poetry to modern Chinese poetry.
Furthermore, existing evaluation criteria are typically designed for ordinary texts or poems with specific restrictions and are not fully applicable to modern poetry. Therefore, we refined these criteria to better capture the nuances of contemporary poetry translation. We engaged a panel of professional poets for assessments and complemented their evaluations with those conducted using GPT-4. The results from both human and machine evaluations demonstrate that our EAPMT method outperforms traditional translation techniques of ChatGPT and existing online systems.


The contributions of our work are as follows:
\begin{itemize}[itemsep=0pt]
    \item We are the first to examine ChatGPT's capabilities in English-Chinese modern poetry translation tasks.
    \item We construct and release a high-quality bilingual poetry dataset.
    \item We identify the optimal prompts and examples (shots) for ChatGPT to effectively translate poetry.
    \item We propose a novel method for poetry translation that uses monolingual poetry explanations as guiding information. This method significantly enhances ChatGPT's performance in translating modern poetry and can be extended to other language pairs and models of language understanding and generation.
    \item We design a new framework for human evaluation criteria specifically applicable to modern poetry translation and engage several professional poets to evaluate the translation results.
\end{itemize}

\section{Related Work}
The earliest work on poetry machine translation employed phrase-based machine translation systems to translate French poetry into metrical English poetry \cite{genzel2010poetic}. This research explored how statistical machine translation systems could produce translations that adhere to constraints such as length, rhythm, and rhyme.
Subsequently, \citet{greene2010automatic} used statistical methods to translate rhymed poetry, achieving the translation of Italian poetry into English. Through unsupervised learning, they identified word stress patterns in an original poetry corpus, which were then utilized in generating English poetry.
\citet{ghazvininejad2018neural} introduced the first neural poetry translation system, capable of translating French poetry into English while adhering to user-specified target rhythm and rhyme patterns. This system demonstrated greater flexibility than phrase-based machine translation (PBMT) systems, consistently generating translations that fit any specified scheme.
\citet{chakrabarty2021don} conducted an empirical study on poetry translation, highlighting a crucial yet often overlooked issue: advanced machine translation systems, trained on large amounts of non-poetry data, struggle to preserve poetic style even if they can maintain meaning and fluency.
In 2022, a Hybrid Machine Translation (HBMT) model was proposed \cite{chakrawarti2022machine}, which improves the semantic and syntactic accuracy of the translation framework. This model effectively translates Hindi poetry into English, demonstrating notable advancements in maintaining the integrity of the original poetic content.

\section{Construction of \DATASET ~Dataset}

Translators need to have a strong foundation in both the source and target languages \cite{GuoMoruo1983GuoMoruoonCreation, wang2006guoview}. Additionally, poetry translators must understand poetry and ideally write poems themselves.
In this paper, we constructed a high-quality bilingual poetry dataset, \DATASET, consisting of 400 modern poems (7,000 lines) in English and their Chinese translations. Our dataset is selected from the high-quality online platform Poetrysky.\footnote{http://poetrysky.com/} We were meticulous in our selection process, ensuring that only high-quality data was included. This rigor is in part due to the expertise of one of our authors, a professional poet who has rigorously filtered out any data of inferior quality.
It is important to note that the judges involved in our human evaluations did not overlap with the translators of the dataset, ensuring unbiased assessments.
The source poems in this dataset were written by multiple professional English poets, with translations from various bilingual poetry translators who are native Chinese speakers. These translators are also accomplished poets, proficient in English, and have extensive experience translating poetry from English to Chinese. Therefore, the translations in this dataset can be considered the gold standard.
The statistics of our dataset are shown in Table \ref{dataset's statistics}. 

\begin{table}
\centering
\begingroup
\begin{tabular}{cccc}
\hline
\textbf{} & {Poems} & {Lines} & {Tokens} \\
\hline
{Datasets} & 400 & 7K & 40K \\
\hline
\end{tabular}
\endgroup
\caption{\label{dataset's statistics}
The statistics of \DATASET ~dataset.
}
\end{table}

\paragraph{Data Validation Experiments}
To address the concerns about the potential exposure of poems in ChatGPT's training data~\cite{shi2023detecting}, we used all the poems (10 poems) in the test set of the EAPMT experiment to conduct data verification experiments on GPT-4\footnote{We used the version gpt-4-1106 of GPT-4 in this study.} to verify the originality of ChatGPT responses. Our data verification experiments consisted of two stages:
\begin{enumerate}[label=\arabic*., itemsep=0pt]
    \item \textit{Poem Continuation}: We prompted the models with the initial 50\%, 70\%, and 90\% of a poem, asking the models to generate the remaining portions. This aimed to verify if the poem had been seen by ChatGPT.
    \item \textit{Translation Continuation}: We applied the same approach to the Chinese translations, providing varying degrees of content (50\%, 70\%, and 90\%) and instructing the models to generate the complete translations.~We then evaluate the similarity against the human translations using SacreBLEU~\cite{post-2018-call}.
\end{enumerate}

\begin{table}
\centering
\begingroup
\begin{tabular}{llll}
\hline
\textbf{} & {50\%} & {70\%} & {90\%} \\
\hline
{Source Poem} & 0.4 & 0.6 & 0.8 \\
{Translation} & 0.3 & 0.5 & 2.5 \\
\hline
\end{tabular}
\endgroup
\caption{\label{SacreBLEU of Data Validation Experiments}
SacreBLEU Scores from data validation experiments
}
\end{table}

As shown in Table \ref{SacreBLEU of Data Validation Experiments}, the outcomes indicated that the model's continuations and translations received low scores in the evaluation, suggesting that the model might not have seen these specific texts. Although we cannot categorically confirm the model's unfamiliarity with the poems, the results highlight areas where model-generated translations could be improved. We refer readers to Appendix \ref{subsec:Examples of the Poem Continuation} for the validation details.





\section{Task Setups}

\subsection{Appropriate Prompts}

The output of ChatGPT depends on the input prompts \cite{antar2023effectiveness}, so it is essential to design effective prompts. Based on the work of \citet{jiao2023chatgpt} and \citet{gao2023design}, and considering the characteristics of modern poetry, we asked GPT-4 for advice and obtained five prompts that can guide it to translate English poetry into modern Chinese poetry at the highest quality. The five prompts advised by GPT-4 are recorded as $\mathcal{P}_{i\in\{1...5\}}$, respectively. More details about GPT-4 providing translation prompts are provided in Appendix \ref{GPT-4 Providing Translation Prompts}.

In addition, we designed three simple poetry translation prompts, recorded as $\mathcal{H}_{i\in\{1...3\}}$, respectively:

\begin{itemize}
\item \textbf{\HP{1}}: \textit{Please provide the Chinese translation for these sentences:}
\item \textbf{\HP{2}}: \textit{Please provide the Chinese translation for this poem:}
\item \textbf{\HP{3}}: \textit{Please translate this English poem into modern Chinese poetry:}
\end{itemize}

To choose the most suitable prompt for modern poetry translation from the above eight prompts, we randomly selected eight poems from the dataset as the test set for the prompt experiment and translated them using GPT-3.5\footnote{We used the gpt-3.5-turbo version in this study.} and GPT-4 with different prompts.

\subsection{Evaluation}

\paragraph{Automatic Evaluation}

Compared with human evaluation, automatic evaluation is widely used because of its lower cost and faster speed. Most natural language generation task evaluations rely on BLEU \cite{novikova2017we}. For our automatic evaluation, we used SacreBLEU \cite{post-2018-call}, BERTScore \cite{zhang2019bertscore}, and COMET \cite{rei-etal-2022-comet} to evaluate the performance of different candidate translations, as BLEU tends to give very low scores. The results are reported in Table \ref{AutoEvaluation}.
From the results of SacreBLEU, it can be seen that \HP{1} achieved the highest scores on both GPT-3.5 and GPT-4, which are at least 3.2 points and 1.5 points higher than the prompts provided by GPT-4, respectively. According to BERTScore and COMET, our designed \HP{2} achieves the highest score on GPT-3.5, while \GP{4} achieves the highest score on GPT-4. However, the differences in scores among the eight prompts within the same model are relatively minor.
Based on these observations, our subsequent experiments were carried out using the designed prompts rather than those generated by the GPT models.

\begin{table*}
\centering
\begingroup
\begin{tabular}{llrrrrrrrr}
\hline
Metrics & Models & \GP{1} & \GP{2} & \GP{3} & \GP{4} & \GP{5} & \HP{1} & \HP{2} & \HP{3} \\
\hline
\multirow{2}{*}{{SacreBLEU}} & {GPT-3.5} & 6.6 & 12.1 & 11.7 & 10.6 & 11.4 & \textbf{15.3} & 15.1 & 14.2 \\
                                   & {GPT-4} & 7.8 & 9.1 & 9.9 & 13.9 & 11.1 & \textbf{15.4} & 14.2 & 13.1 \\
\hline
\multirow{2}{*}{{BERTScore}} & {GPT-3.5} & 0.5603 & 0.5648 & 0.5658 & 0.5601 & 0.5616 & 0.5766 & \textbf{0.5777} & 0.5728 \\
                                   & {GPT-4} & 0.5455 & 0.5583 & 0.5533 & \textbf{0.5721} & 0.5589 & 0.5686 & 0.5693 & 0.5662 \\
\hline
\multirow{2}{*}{{COMET}} & {GPT-3.5} & 0.7750 & 0.7895 & 0.7865 & 0.7746 & 0.7816 & 0.7891 & \textbf{0.7991} & 0.7917 \\
                                & {GPT-4} & 0.7636 & 0.7677 & 0.7627 & \textbf{0.7876} & 0.7732 & 0.7812 & 0.7815 & 0.7774 \\
\hline
\end{tabular}
\endgroup
\caption{\label{AutoEvaluation}
Results of English poetry translated into modern Chinese poetry under different prompts, evaluated with SacreBLEU, BERTScore (F1), and COMET metrics. $\mathcal{P}_{i\in\{1...5\}}$ are the five prompts provided by GPT-4 itself, while $\mathcal{H}_{i\in\{1...3\}}$ are the prompts we designed.
}
\end{table*}






\paragraph{Human Evaluation}
Automatic evaluation metrics are not always appropriate for natural language generation, especially for poetry-related tasks \cite{novikova2017we, refaee2023okaz}, where the translation of poetry cannot be directly evaluated due to its unique characteristics \cite{GuoMoruo, GuoMoruo1983GuoMoruoonCreation}. The best poetry translators are often both poets and translators \cite{GuoMoruo1983GuoMoruoonCreation, wang2006guoview}. Therefore, we invited six judges to evaluate the experimental results of eight modern English poems translated under eight different prompts. These judges are Chinese professional poets born between the 1960s and 1990s, who are members of the Chinese Writers Association.\footnote{https://www.chinawriter.com.cn} They have a high level of poetry creation among poets of their age group and possess their own representative poems or poetry collections. These judges represent different eras in Chinese poetry. Among them are university professors specializing in literature, PhDs in poetry, and professional translators of English-Chinese poetry. Importantly, they are native Chinese speakers who understand English, with some having engaged in poetry writing or English-Chinese poetry translation for many years.
Specifically, we asked the judges to vote on the output translations of GPT-3.5 and GPT-4 under three human prompts through a questionnaire survey to select the most appropriate prompt. For three translations of the same source language poem, each judge had only one vote to choose the poem they believed to be of the best quality. Thus, each judge had a total of eight votes for GPT-3.5 and GPT-4, respectively. To obtain an objective and fair evaluation and avoid any bias against artificial intelligence or large models, we did not inform the judges that the poems were translated by ChatGPT when we invited them to complete the questionnaire.
Table \ref{Voting results of prompts} shows the voting results of the six professional judges on the output translations of eight poems by GPT-3.5 and GPT-4 under different prompts.
For GPT-3.5, \HP{3} received 21 votes, at least 5 votes more than the other two prompts. For GPT-4, \HP{2} received the most votes. Ultimately, we concluded that for GPT-3.5, the best prompt is \HP{3}, while for GPT-4, the best prompt is \HP{2}.

\begin{table}
\centering
\begingroup
\begin{tabular}{llll}
\hline
\textbf{} & \HP{1} & \HP{2} & \HP{3} \\
\hline
{GPT-3.5} & 11 & 16 & \textbf{21} \\
{GPT-4} & 14 & \textbf{19} & 15 \\
\hline
\end{tabular}
\endgroup
\caption{\label{Voting results of prompts}
Voting results of six professional judges on the output translation of eight poems by GPT-3.5 and GPT-4 under different prompts.
}
\end{table}

The results from both automatic and human evaluation demonstrate that the prompts we crafted achieve better results in poetry translation tasks. Compared to other prompts, \HP{3} is the most appropriate for GPT-3.5, whereas \HP{2} is the most appropriate for GPT-4.

\subsection{Appropriate Demonstrations}
By exploring the translation performance of GPT-3.5 and GPT-4 on modern poetry in different small sample scenarios, we aim to find the most appropriate method. Specifically, we randomly selected another ten poems (with no overlap with the poems in previous experiments) from the dataset as a new test set. GPT-3.5 and GPT-4 were tested under $\{0,1,3,5\}$-shot scenarios in sequence using the most appropriate prompts identified from previous experiments. To ensure the fairness and reliability of the experiment, the demonstration examples used were excluded from the test samples \cite{gao2023design}. The prompts we designed for few-shot scenarios are provided in Appendix \ref{subsec:Few-shot Prompt}.

\begin{table}
\centering
\begingroup
\begin{tabular}{lcccc}
\hline
\textbf{} & {0-shot} & {1-shot} & {3-shot} & {5-shot} \\
\hline
{GPT-3.5} & \textbf{25} & 13 & 12 & 10 \\
{GPT-4} & \textbf{19} & 16 & 13 & 12 \\
\hline
\end{tabular}
\endgroup
\caption{\label{Voting results of shot}
Voting results on the output translation of ten poems in different shot scenarios.
}
\end{table}

\paragraph{Evaluation}
We invited an additional six professional poets to evaluate the model's outputs through a questionnaire, aiming to select the highest-quality translations under different shot scenarios. For translations of the same poem in different cases, each judge had only one vote to choose the poem they deemed to be of the best quality. Thus, each judge had a total of ten votes for GPT-3.5 and GPT-4, respectively. To ensure an unbiased evaluation, the judges were not informed of the origins of the different poems when casting their votes. Table \ref{Voting results of shot} presents the voting results.
Both GPT-3.5 and GPT-4 received the most votes in the 0-shot scenario, which is contrary to the findings of \citet{gao2023design}. Unexpectedly, as the number of samples increases, the performance of the models decreases. This suggests that the conclusions of previous work are not applicable to modern poetry translation. More samples did not promote the models to learn the characteristics of poetry.
The superior performance in the zero-shot scenario is likely due to the inherent nature of modern poetry, which is characterized by its freedom and lack of constraints \cite{awan2015new}. The essence of modern poetry lies in its open forms, free genres, departure from traditional narratives, and the absence of fixed language patterns \cite{GuoMoruo, wang2006guoview, skerratt2013form}. Providing demonstrations in the prompts appears to curtail the model's capacity for generating diverse translations, thereby constraining the inherently flexible nature of modern poetry. This is evidenced by the results presented in Table \ref{Voting results of shot}, where an increase in provided examples inversely correlates with human votes.
Based on these observations, we can draw the following conclusions: The optimum performance of GPT-3.5 on the poetry translation task occurs in the 0-shot scenario using the third prompt (\HP{3}), and the best translation result is denoted as GPT3.5-Best. The optimum performance of GPT-4 occurs in the 0-shot scenario using the second prompt (\HP{2}), and the best translation result is recorded as GPT4-Best.

\subsection{Case Study}

Table \ref{Examples of Poem Translation} of Appendix \ref{subsec:Examples of Poem Translation} presents the best translations of GPT-3.5 and GPT-4 on the poem, ``Balance.''\footnote{http://poetrysky.com/quarterly/quarterly-4-carrigan.html}



\paragraph{Not Accurate Enough} Every word in the poem cannot be replaced by another word, and it is necessary to find the most suitable translation \cite{owen1990world}. However, as we can see from the examples, GPT-3.5 and GPT-4 typically translate based on the surface meaning of the text, rather than choosing the most appropriate words based on context. For instance, ``anchors'' is translated literally as ``\begin{CJK}{UTF8}{gkai}锚定\end{CJK} (anchor)'' by GPT-3.5 and ``\begin{CJK}{UTF8}{gkai}稳固\end{CJK} (stable)'' by GPT-4. While ``\begin{CJK}{UTF8}{gkai}稳固\end{CJK} (stable)'' is more suitable than ``\begin{CJK}{UTF8}{gkai}锚定\end{CJK} (anchor),'' it still does not conform to the context of the original poem and is far from the meaning of ``\begin{CJK}{UTF8}{gkai}停泊\end{CJK} (anchor)'' in the human translation. Additionally, ``applauds'' is translated by GPT-3.5 and GPT-4 as ``\begin{CJK}{UTF8}{gkai}拍手欢呼\end{CJK} (clapping and cheering)'' and ``\begin{CJK}{UTF8}{gkai}鼓掌\end{CJK} (handclap),'' respectively, losing the literary meaning.



\paragraph{Room for Improvement in Line-breaking} Good poetry cannot exist without appropriate line-breaking. Line-breaking is more profound than the poem itself; it not only creates internal tension within the poem but also allows the reader to appreciate its beauty at a glance \cite{xue2016something}. However, there is still significant room for improvement in the line-breaking of translations generated by GPT-3.5 and GPT-4. For example, the sentence ``She felt water on its ordained/journey to the sea'' in the source poem was translated into a grammatically accurate sentence by GPT-4, resulting in the disappearance of tension between the lines of the poem. When translating English into Chinese, grammatical rules should not strictly be followed \cite{chen1993language}; the narrative can be incoherent, and the combination of words is free \cite{awan2015new}.

\paragraph{Poeticity Is Not Well-perserved} The poeticity of the source poem should be preserved in translation \cite{GuoMoruo, bianzhilin1989translation, skerratt2013form}. However, we found that the most poetic sentence in the source poem, ``pass beneath her rippling image,'' was lost when translated literally by GPT-3.5 and GPT-4. The original poem and the reference translation use anthropomorphic rhetoric to compare ``water'' to human beings, so it is translated as ``\begin{CJK}{UTF8}{gkai}经过\end{CJK}.'' However, GPT-3.5 and GPT-4 translate ``pass'' as ``\begin{CJK}{UTF8}{gkai}流过\end{CJK},'' which clearly does not capture the meaning of the source poem, resulting in the omission of the poeticity during the translation process.

\section{Explanation-Assisted Poetry Machine Translation}

In light of the experimental results and analysis presented, it is evident that while ChatGPT can interpret and analyze poems \cite{chen3using}, it still encounters numerous challenges. To enhance the performance of ChatGPT in poetry translation, we introduce a novel approach termed Explanation-Assisted Poetry Machine Translation (EAPMT). This method leverages monolingual poetry explanations as guidance information.
Specifically, the process begins with ChatGPT generating an explanation for a poem presented in the source language. This explanation encompasses both the literal content and deeper meanings, contextualized within the cultural backdrop of the poem. Subsequently, this explanation serves as guidance for ChatGPT to produce a translation into the target language. The two-step framework ensures that nuances and subtler aspects of the poem, often lost in conventional translation processes, are preserved.
Figure \ref{EAPMT} contrasts the novel Explanation-Assisted Poetry Machine Translation (EAPMT) framework with traditional translation methods, highlighting the advantages of the EAPMT approach.
Table \ref{EAPMT prompt} presents the prompts designed for the translation process.
A complete example of the EAPMT method is illustrated in Appendix \ref{Examples of the EAPMT method}.


\begin{table}[H]
\centering
\renewcommand{\arraystretch}{0.9}
\begin{tabular}{p{7cm}}
\hline
\MYFONTSIZE{\textbf{GPT-3.5:}}  \\
\MYFONTSIZE{Step 1}  \\
\MYFONTSIZE{Please provide an explanation for this English poem:}  \\
\MYFONTSIZE{English poem:\{\{$X_i$\}\}}  \\
\MYFONTSIZE{Explanation:\{\{$M_i$\}\}}  \\
\vspace{0.01cm}
\MYFONTSIZE{Step 2}  \\
\MYFONTSIZE{Please translate this English poem into a modern Chinese poem based on its explanation:}  \\
\MYFONTSIZE{Explanation:\{\{$M_i$\}\}}  \\
\MYFONTSIZE{English poem:\{\{$X_i$\}\}}  \\
\MYFONTSIZE{Modern Chinese poem:}  \\
\hline
\MYFONTSIZE{\textbf{GPT-4:}}  \\
\MYFONTSIZE{Step 1}  \\
\MYFONTSIZE{Please provide an explanation for this poem:}  \\
\MYFONTSIZE{Poem:\{\{$X_i$\}\}} \\
\MYFONTSIZE{Explanation:\{\{$M_i$\}\}}  \\
\vspace{0.01cm}
\MYFONTSIZE{Step 2} \\
\MYFONTSIZE{Please provide the Chinese translation for this poem based on its explanation:}  \\
\MYFONTSIZE{Explanation:\{\{$M_i$\}\}} \\
\MYFONTSIZE{Poem:\{\{$X_i$\}\}}  \\
\MYFONTSIZE{Chinese translation:}  \\

\hline

\end{tabular}
\renewcommand{\arraystretch}{1.0}
\caption{\label{EAPMT prompt}
The prompts designed for Explanation-Assisted Poetry Machine Translation.
}
\end{table}

\section{Experiments}
Our experiments were performed using GPT-3.5 (gpt-3.5-turbo) and GPT-4 (gpt-4-1106), and involved comparisons with leading online translation systems. For Chinese literary texts, Baidu Translate\footnote{https://fanyi.baidu.com} demonstrated superior performance over Google Translate \cite{M2019Exploring}. Consequently, Baidu Translate was selected as the online translation system for our study. Table \ref{Examples of Poem Translation} of Appendix \ref{subsec:Examples of Poem Translation} presents the translations of our EAPMT method on the poem, ``Balance.''

\section{Evaluation}

\subsection{Human Evaluation}
Previous work on poetry generation has utilized diverse human evaluation criteria such as fluency, coherence, meaningfulness, poeticness, overall quality, and adequacy \cite{Manurung2012UsingGA, zhang2014chinese, yan2016, yi2017generating, yi2018automatic, Seljan2020HumanQE, refaee2023okaz}. However, when translating English into Chinese, strict adherence to grammatical rules is not always necessary \cite{chen1993language, owen2003stepping}, and modern poetry often features discontinuous narration and a flexible combination of words \cite{awan2015new}. Thus, our human evaluation framework does not prioritize fluency and coherence.
Given the unique and irreplaceable nature of each word in a poem \cite{owen1990world}, accuracy remains an indispensable criterion in our evaluation of poetry translation. Additionally, the aspect of line-breaking, which distinguishes modern from classical poetry, plays a critical role in conveying the intrinsic rhythm and creating inner tension, thereby enhancing the visual and architectural beauty of the poem \cite{wang2018review, xue2016something}. For these reasons, appropriate line-breaking is also included in our evaluation criteria.
Drawing from previous studies \cite{Manurung2012UsingGA, zhang2014chinese, yan2016, yi2017generating, yi2018automatic, Seljan2020HumanQE, refaee2023okaz} and reflecting the unique elements of modern poetry, we have designed a new human evaluation framework specifically tailored for the translation of modern poetry. We assess candidate translations comprehensively, focusing on eight key aspects, ranging from overall impact to specific details:

\begin{itemize}[itemsep=0pt]
    \item \textbf{Overall Impression (OI)}: This criterion evaluates the general impact of the candidate translation as compared to the source poem or reference translation. It assesses whether the translation successfully captures the essence and tone of the original.

    \item \textbf{Similarity (Sim)}: Measures the degree of similarity between the candidate translation and the reference translation, focusing on stylistic and thematic alignment.
    
    \item \textbf{Fidelity (Fide)}: Assesses how faithfully the translation conveys the original poem's intent, emotions, and deeper meanings, thus evaluating whether the translation transcends mere linguistic equivalence to preserve the poem's core essence.
    
    \item \textbf{Line-breaking (Line)}: Evaluates the appropriateness of line breaks in the translation in comparison to the source poem or reference translation, considering how these contribute to the poem's rhythm and tension.
    
    \item \textbf{Meaningfulness (Mean)}: Examines the extent to which the translation conveys the original poem's meanings, exploring both surface-level content and deeper interpretative layers.
    
    \item \textbf{Poeticity (Poet)}: Assesses how well the poetic qualities of the original text, such as imagery, metaphor, and overall poetic effect, are preserved in the translation.
    
    \item \textbf{Accuracy (Acc)}: Focuses on the precision of translated elements, including words and word combinations, crucial to maintaining the integrity of the poem.
    
    \item \textbf{Errors (Erro)}: Identifies and categorizes errors in the translation, with a detailed scoring system that ranges from minor, ignorable mistakes to significant errors that alter the poem's meaning.
\end{itemize}
\vspace{-0.01cm}
Each of these criteria utilizes a 6-point rating scale, where a score of 5 indicates equivalence to the reference translation, and a score of 6 signifies a superior quality compared to the reference translation.
To ensure an objective evaluation, we enlisted six professional judges who were provided with only the source poem and the reference translation. The identities of the translators (human or machine) behind the candidate translations were not disclosed to the judges.
Table \ref{human evaluation results} presents the human evaluation results, comparing our proposed method with traditional translation methods and online translation systems, thereby elucidating the effectiveness of our approach in translating modern poetry.

\begin{table*}
\centering

\begingroup
\begin{tabular}{lccccccccc}  
\hline
\textbf{} & {OI} & {Sim} & {Fide} & {Line} & {Mean} & {Poet} & {Acc} & {Erro} \\
\hline
{Online System} & 3.00 & 3.02 & 3.15 & 3.85 & 3.27 & 3.08 & 3.12 & 3.00 & \textbf{} \\
\hdashline
{GPT3.5-Best} & 3.58 & 3.50 & 3.60 & 4.50 & 4.02 & 3.82 & 3.75 & 3.70 & \textbf{} \\
{EAPMT-3.5} & \textbf{3.82} & \textbf{3.58} & \textbf{3.67} & \textbf{4.70} & \textbf{4.08} & {3.93} & \textbf{4.10} & \textbf{3.80} & \textbf{} \\
\hdashline
{GPT4-Best} & 3.77 & 3.53 & 3.65 & 4.25 & 3.97 & 3.82 & 3.82 & 3.70 & \textbf{} \\
{EAPMT-4.0} & \textbf{4.00} & \textbf{3.60} & \textbf{3.80} & \textbf{4.58} & \textbf{4.15} & \textbf{4.05} & \textbf{4.13} & \textbf{3.87} & \textbf{} \\
\hline
\end{tabular}
\endgroup
\caption{\label{human evaluation results}
Results from human evaluations comparing our proposed method with traditional approaches and online translation systems.
}
\end{table*}

\section{Analysis}

As shown in Table \ref{human evaluation results}, both our proposed method and traditional methods using GPT-3.5 and GPT-4 outperform online systems across all evaluation criteria.
Furthermore, under all evaluation criteria, our method scored higher than the best performances of various versions of ChatGPT. Notably, in terms of accuracy, line-breaking, and overall impression, our EAPMT method achieved significant improvements, with scores at least 0.20 points higher than those of traditional methods (GPT3.5-Best \& GPT4-Best).
Traditional machine translation faces two primary challenges: first, addressing the ambiguities arising from the different ways languages express the same concepts \cite{andrabi2021review}; second, handling poor translation results when encountering rare words \cite{arthur2016incorporating, sennrich2015neural, luong2014addressing}.
Our method effectively mitigates these issues in traditional machine translation. Specifically, in terms of accuracy, our method surpasses traditional methods by 0.35 points on GPT-3.5 and 0.31 points on GPT-4. This improvement is attributed to GPT-4's ability to generate poetic explanations based directly on the source poem, capturing its exact meaning and appropriate emotions, and even replicating specific words from the source. By incorporating these explanations into the prompts as guidance, GPT models can more accurately grasp the meaning of the source poem, thus reducing issues related to ambiguity and polysemy in translation.
In terms of line-breaking, our method shows a 0.20-point increase with GPT-3.5 and a 0.33-point increase with GPT-4 compared to traditional methods. This is because the source poem in the prompt includes explicit line-breaking information, enabling ChatGPT to better learn and apply this information while comprehending the specific meaning of the poem.
Regarding overall impression, our method achieved scores of 0.24 and 0.23 points higher than traditional methods on GPT-3.5 and GPT-4, respectively. This aligns with our survey results, where we invited 10 poets to evaluate what factors influenced their judgment of overall impression. All poets indicated that poeticity was the most critical factor, followed by line-breaking.

\begin{table}
\centering
\begin{tabular}{lcc}
\hline
\textbf{} & {Line} & {Acc} \\
\hline
{EAPMT-3.5} & 3 & 7 \\
{EAPMT-4.0} & 3 & 5 \\
\hline
\end{tabular}
\caption{\label{6-score}
Count of each criterion receiving a score of 6 in the evaluation.
}
\end{table}

Surprisingly, we observed a score of 6 in the human evaluation results. To further analyze this, we counted the instances where each evaluation criterion received a score of 6 across six human judges evaluating the candidate translations of ten poems. As illustrated in Table \ref{6-score}, our proposed method achieved a score of 6 for both line-breaking and accuracy when using GPT-3.5 and GPT-4, indicating superior performance compared to the reference translations in these criteria. Notably, in terms of accuracy, our method was awarded a score of 6 on seven occasions with GPT-3.5 and five occasions with GPT-4.

\subsection{Evaluation Using GPT-4}

Previous studies, such as those by \citet{chen3using}, have demonstrated that models like ChatGPT can effectively analyze and interpret the meaning of poetry. Building on this foundation, we employed GPT-4 to evaluate various translations based on the human evaluation criteria we established. As detailed in Appendix \ref{subsec:GPT-4 evaluation details}, the evaluation results for GPT-4 indicate that our method surpasses traditional approaches in terms of overall impression, fidelity, meaningfulness, and poeticity. These findings reaffirm that our method significantly enhances the performance of GPT-4 in poetry translation tasks, as shown in Table \ref{GPT-4 Evaluation Prompts} in Appendix \ref{subsec:GPT-4 evaluation details}.


\paragraph{Determining Poeticity in Translated Texts}
Poeticity is inherently subjective; however, within the lines of the reference translation, one distinctly stands out as the most poetic. To explore whether GPT-3.5 and GPT-4 can accurately assess poetic qualities, we engaged six professional poets to select the most poetic sentence from each of ten poems. The sentence receiving the majority of votes was considered the most poetic. Concurrently, we tasked GPT-3.5 and GPT-4 with identifying the most poetic sentence from each of the ten source poems in English and their corresponding translations into Chinese, using language-specific prompts. We then evaluated the consistency of the selections made by GPT-3.5 and GPT-4 against those chosen by the human poets.
As illustrated in Table \ref{the most poetic sentences}, GPT-3.5 correctly identified the most poetic sentences in five English poems and two Chinese poems, indicating significant potential in recognizing poetic elements within English poetry. In contrast, GPT-4 accurately identified the most poetic lines in two English poems and three Chinese poems, demonstrating superior comprehension of Chinese poetry compared to GPT-3.5. These findings underscore the effectiveness of our designed prompts and the reliability of the human evaluation results. Specifically, in HP2, which exclusively involved Chinese poetry, human evaluation results indicated that \HP{2} optimally harnesses GPT-4's potential in poetry translation. Similarly, \HP{3}, which involved translating English poetry into Chinese, showed that HP3 maximizes GPT-3.5's capabilities in poetry translation.
This study not only demonstrates the varying capabilities of GPT models in identifying poetic qualities across languages but also highlights the importance of tailored prompt design to optimize model performance in specific linguistic contexts.

\begin{table}
\centering
\begin{tabular}{lcc}
\hline
\textbf{} & {English} & {Chinese} \\
\hline
{GPT-3.5} & \textbf{5} & 2 \\
{GPT-4} & 2 & \textbf{3} \\
\hline
\end{tabular}
\caption{\label{the most poetic sentences}
Consistency of GPT-3.5 and GPT-4 in identifying the most poetic sentences from ten poems compared to human judgments.
}
\end{table}



\section{Conclusion}
In this work, we examine ChatGPT's capabilities in English-Chinese poetry translation tasks by utilizing targeted prompts and small sample scenarios to ascertain optimal performance. To address ChatGPT's shortcomings, we propose an Explanation-Assisted Poetry Machine Translation (EAPMT) method, which leverages monolingual poetry explanation as a guiding information for the translation process. Furthermore, we refine existing evaluation criteria to better suit the nuances of contemporary poetry translation. The results from both human and machine evaluations demonstrate the effectiveness of our EAPMT method.
In future work, we will try to correct the explanation generated in the first step with human experts or ChatGPT, thereby further improving the accuracy of the translation.

\section*{Limitations}
Our research focuses on modern poetry, which has a very free format \cite{skerratt2013form, awan2015new, duanjinyan2008research}. Our method may not be applicable to classical poetry and rhyming modern poetry. Therefore, the results of our study cannot represent or cover all types of poetry.


\section*{Ethics Statement}
\begin{itemize}
    \item \textbf{Datasets} The dataset we built consists of high-quality poems and translations. Poetry may contain negative emotions, but there is no information harmful to society.
    \item \textbf{Prejudice} In order to ensure the fairness and reliability of the results, we did not tell the human judges that the content they evaluated was generated by artificial intelligence during the process of human evaluation, but we told the truth to the judges after the evaluation and won the support of the judges.
    \item \textbf{The impact of this work on human translators} After the human evaluation was completed, we held discussions with professional poetry translators who work across various language pairs including English-Chinese, Russian-Chinese, Portuguese-English-Chinese, and Japanese-English-Chinese. They all concur that artificial intelligence, while a significant advancement, poses no existential threat to the domain of poetry translation. This consensus stems from the belief that AI technologies, including our own contributions, have not yet achieved the nuanced comprehension of modern poetry that human expertise offers. Conversely, these translators view our work as a promising support tool for novices in poetry translation, facilitating their initial forays into this intricate field.
    
\end{itemize}

\section*{Acknowledgements}
This work was supported in part by the Science and Technology Development Fund, Macau SAR (Grant No. FDCT/060/2022/AFJ, the mainland China collaboration project, National Natural Science Foundation of China Grant No. 62261160648), the Science and Technology Development Fund, Macau SAR (Grant No. FDCT/0070/2022/AMJ, the mainland China collaboration project, China Strategic Scientific and Technological Innovation Cooperation Project Grant No. 2022YFE0204900), the Multi-year Research Grant from the University of Macau (Grant No. MYRG-GRG2023-00006-FST-UMDF), and the Tencent AI Lab Rhino-Bird Gift Fund (Grant No. EF2023-00151-FST). 

\bibliography{custom}

\clearpage
\appendix
\section{Appendix}
\label{sec:appendix}

\subsection{Validation Experiments}
\label{subsec:Examples of the Poem Continuation}

To validate if the poems have been seen by the models, we prompted the models by providing various portions of the poem (or target translation) and asked the models to generate the rest of it. Table \ref{Validation_Prompt} shows an example of Poem Continuation using GPT-4 to complete an English poem, “Balance,” by providing the first 50\% of the poem's context.




\begin{table*}
\centering
\renewcommand{\arraystretch}{0.9}
\begin{tabular}{p{15cm}}
\hline
\MYFONTSIZE{\textbf{Source Poem: Balance}}  \\
\vspace{-0.005cm}
\MYFONTSIZE {The white crane}  \\
\MYFONTSIZE{anchors to her liquid mirror} \\
\MYFONTSIZE{on one improbable leg.}  \\
\MYFONTSIZE{She feels water on its ordained}  \\
\MYFONTSIZE{journey to the sea}  \\
\MYFONTSIZE{{pass beneath her rippling image.}}  \\
\MYFONTSIZE{{It is, she knows,}}  \\
\MYFONTSIZE{{the way of things.}}  \\
\MYFONTSIZE{{On the shore, bamboo}}  \\
\MYFONTSIZE{{applauds}}  \\
\MYFONTSIZE{{such grace}}  \\
\MYFONTSIZE{{with swaying hands}}  \\
\MYFONTSIZE{{clapping in the wind.}}  \\
\hline
\MYFONTSIZE{\textbf{Input Prompt}}  \\
\vspace{-0.005cm}
\MYFONTSIZE{{Please continue writing the next 6 lines of the modern poem entitled ``Balance'', which requires a total of 13 lines:}}  \\
\MYFONTSIZE{{The white crane}}  \\
\MYFONTSIZE{{anchors to her liquid mirror}}  \\
\MYFONTSIZE{{on one improbable leg.}}  \\
\MYFONTSIZE{{She feels water on its ordained}}  \\
\MYFONTSIZE{{journey to the sea}}  \\
\MYFONTSIZE{{pass beneath her rippling image.}}  \\
\MYFONTSIZE{{anchors to her liquid mirror}}  \\
\MYFONTSIZE{{It is, she knows,}}  \\
\hline
\MYFONTSIZE{\textbf{Model Output}}  \\
\vspace{-0.005cm}
\MYFONTSIZE{{a delicate dance of give and take,}}  \\
\MYFONTSIZE{{the art of holding still while letting go.}}  \\
\MYFONTSIZE{{Around her, life in ceaseless flux,}}  \\
\MYFONTSIZE{{yet she centers herself in the flow,}}  \\
\MYFONTSIZE{{a serene pivot in the world's relentless churn,}}  \\
\MYFONTSIZE{{balancing the simple with the sublime.}}  \\

\hline

\end{tabular}
\renewcommand{\arraystretch}{1.0}
\caption{\label{Validation_Prompt}
This illustrates the continuation of the English poem ``Balance'' using GPT-4, which was provided with the first 50\% of the poem as context. The lower part displays the six sentences generated by the model. Notably, there is a significant divergence between GPT-4's continuation and the original poem's latter half.
}
\end{table*}

    

    

    


\subsection{GPT-4 Generated Prompts}
\label{GPT-4 Providing Translation Prompts}

To identify the best candidate prompts, we asked GPT-4 to generate five prompts for translating English poetry into modern Chinese poetry at the highest quality by using the following prompt: \textit{“Provide five concise prompts or templates that can make you translate English poetry into modern Chinese poetry at the highest quality.”}

The prompts advised by GPT-4 are recorded as $\mathcal{P}_{i\in\{1...5\}}$, respectively:
\begin{itemize}
    \item \textbf{$\mathcal{P}_1$}: \textit{Please translate the following modern English poem into modern Chinese poetry, considering its cultural and historical context in which it was written. Maintain the tone, style, and emotional impact of the original poem.}
   \item \textbf{$\mathcal{P}_2$}: \textit{Translate this modern English poem into modern Chinese poetry, focusing on preserving the vivid imagery and metaphoric language. Ensure the translation conveys the same visual and sensory experiences as the original.}
   \item \textbf{$\mathcal{P}_3$}: \textit{Translate the following modern English poem into modern Chinese poetry, making sure to maintain its rhyme scheme and rhythm to the best extent possible. The translation should attempt to replicate the musicality and flow of the original text.}
   \item \textbf{$\mathcal{P}_4$}: \textit{Translate this modern English poem into modern Chinese poetry, ensuring the literal meaning of each line is accurately conveyed. The focus here is on the direct translation of the words and phrases, rather than on preserving the poetic devices used in the original.}
   \item \textbf{$\mathcal{P}_5$}: \textit{Translate the following modern English poem into modern Chinese poetry, taking into account the unique style of the poet. Try to capture the author's voice, style, and idiosyncrasies in the translated version.}
\end{itemize}

\subsection{Few-shot Prompt}
\label{subsec:Few-shot Prompt}


Table \ref{Few-shot prompt} provides the prompt template utilized for conducting few-shot translation experiments, illustrating the structured input provided to the model.

\begin{table*}
\centering
\renewcommand{\arraystretch}{0.9}
\begin{tabular}{p{15cm}}
\hline
\MYFONTSIZE{\textbf{GPT-3.5}} \\
\vspace{-0.005cm}
\MYFONTSIZE{{Please translate this English poem into modern Chinese poetry:}}  \\
\MYFONTSIZE{{Example(s):}}  \\
\MYFONTSIZE{{English Poem:\{\{$X_1$\}\}}}  \\
\MYFONTSIZE{{Modern Chinese Poem:\{\{$Y_1$\}\}}}  \\
\MYFONTSIZE{{...}}  \\
\MYFONTSIZE{{English Poem:\{\{$X_k$\}\}}}  \\
\MYFONTSIZE{{Modern Chinese poem:\{\{$Y_k$\}\}}}  \\
\MYFONTSIZE{{English Poem:\{\{English Poem\}\}}}  \\
\MYFONTSIZE{{Modern Chinese Poem:}}  \\
\hline
\MYFONTSIZE{\textbf{GPT-4}}  \\
\vspace{-0.005cm}
\MYFONTSIZE{{Please provide the Chinese translation for this poem:}}  \\
\MYFONTSIZE{{Example(s):}}  \\
\MYFONTSIZE{{Poem:\{\{$X_1$\}\}}}  \\
\MYFONTSIZE{{Chinese Translation:\{\{$Y_1$\}\}}}  \\
\MYFONTSIZE{{...}}  \\
\MYFONTSIZE{{Poem:\{\{$X_k$\}\}}}  \\
\MYFONTSIZE{{Chinese Translation:\{\{$Y_k$\}\}}}  \\
\MYFONTSIZE{{Poem:\{\{English Poem\}\}}}  \\
\MYFONTSIZE{{Chinese Translation:}}  \\
\hline

\end{tabular}
\renewcommand{\arraystretch}{1.0}
\caption{\label{Few-shot prompt}
This table presents the prompts specifically crafted for our few-shot learning experiments, showcasing the tailored inputs used to optimize model performance in minimal data scenarios.
}
\end{table*}



\subsection{Examples of Poem Translation}
\label{subsec:Examples of Poem Translation}
Table \ref{Examples of Poem Translation} shows an example of poem translation using the best prompts for GPT-3.5 and GPT-4 on the poem, ``Balance.'' The includes the original English poem, its Chinese translation, the translations produced by the GPT models based on their corresponding optimum prompts, as well as the translations generated by our Explanation-Assisted Poetry Machine Translation (EAPMT) method.

\begin{table*}
\centering
\renewcommand{\arraystretch}{0.9}
\begin{tabular}{p{7cm}ll}
\hline
\MYFONTSIZE{\textbf{Source Poem}}  & \MYFONTSIZE{\textbf{Reference Translation}} \\
\MYFONTSIZE{Balance}  & \MYFONTSIZE{\begin{CJK}{UTF8}{gkai}平衡\end{CJK}} \\
\vspace{0.01cm}   & \vspace{0.01cm}\\ 
\MYFONTSIZE{The white crane}  & \MYFONTSIZE{\begin{CJK}{UTF8}{gkai}白鹤\end{CJK}} \\
\MYFONTSIZE{anchors to her liquid mirror}  & \MYFONTSIZE{\begin{CJK}{UTF8}{gkai}停泊在她流体的镜子里\end{CJK}} \\
\MYFONTSIZE{on one improbable leg.}  & \MYFONTSIZE{\begin{CJK}{UTF8}{gkai}以一只不太可能的腿。\end{CJK}} \\
\MYFONTSIZE{She feels water on its ordained}  & \MYFONTSIZE{\begin{CJK}{UTF8}{gkai}她感到水在它注定的\end{CJK}} \\
\MYFONTSIZE{journey to the sea}  & \MYFONTSIZE{\begin{CJK}{UTF8}{gkai}流向海洋的旅程上\end{CJK}} \\
\MYFONTSIZE{pass beneath her rippling image.}  & \MYFONTSIZE{\begin{CJK}{UTF8}{gkai}从她起皱的影子下经过。\end{CJK}} \\
\MYFONTSIZE{It is, she knows,}  & \MYFONTSIZE{\begin{CJK}{UTF8}{gkai}事情，她知道，\end{CJK}} \\
\MYFONTSIZE{the way of things.}  & \MYFONTSIZE{\begin{CJK}{UTF8}{gkai}就是这样。\end{CJK}} \\
\MYFONTSIZE{On the shore, bamboo}  & \MYFONTSIZE{\begin{CJK}{UTF8}{gkai}岸上，竹子们\end{CJK}} \\
\MYFONTSIZE{applauds}  & \MYFONTSIZE{\begin{CJK}{UTF8}{gkai}为如斯的优美\end{CJK}} \\
\MYFONTSIZE{such grace}  & \MYFONTSIZE{\begin{CJK}{UTF8}{gkai}喝彩\end{CJK}} \\
\MYFONTSIZE{with swaying hands}  & \MYFONTSIZE{\begin{CJK}{UTF8}{gkai}用摆动的手\end{CJK}} \\
\MYFONTSIZE{clapping in the wind.}  & \MYFONTSIZE{\begin{CJK}{UTF8}{gkai}在风中频频鼓掌。\end{CJK}} \\
\hline
\MYFONTSIZE{\textbf{GPT3.5-Best Translation}}  & \MYFONTSIZE{\textbf{GPT4-Best Translation}} \\
\MYFONTSIZE{\begin{CJK}{UTF8}{gkai}平衡\end{CJK}}  & \MYFONTSIZE{\begin{CJK}{UTF8}{gkai}平衡\end{CJK}} \\
\vspace{0.01cm}   & \vspace{0.01cm}\\
\MYFONTSIZE{\begin{CJK}{UTF8}{gkai}白鹤\end{CJK}}  & \MYFONTSIZE{\begin{CJK}{UTF8}{gkai}白鹤\end{CJK}} \\
\MYFONTSIZE{\begin{CJK}{UTF8}{gkai}用一条匪夷所思的腿\end{CJK}}  & \MYFONTSIZE{\begin{CJK}{UTF8}{gkai}以一只难以置信的脚\end{CJK}} \\
\MYFONTSIZE{\begin{CJK}{UTF8}{gkai}锚定在她的液态镜面上。\end{CJK}}  & \MYFONTSIZE{\begin{CJK}{UTF8}{gkai}稳固于她的水面镜像。\end{CJK}} \\
\MYFONTSIZE{\begin{CJK}{UTF8}{gkai}她感受着水\end{CJK}}  & \MYFONTSIZE{\begin{CJK}{UTF8}{gkai}她感受到水正按其注定\end{CJK}} \\
\MYFONTSIZE{\begin{CJK}{UTF8}{gkai}在它注定的旅程中\end{CJK}}  & \MYFONTSIZE{\begin{CJK}{UTF8}{gkai}的旅程向大海流去，\end{CJK}} \\
\MYFONTSIZE{\begin{CJK}{UTF8}{gkai}在她涟漪的映像下流过。\end{CJK}}  & \MYFONTSIZE{\begin{CJK}{UTF8}{gkai}流过她波动的倩影下。\end{CJK}} \\
\MYFONTSIZE{\begin{CJK}{UTF8}{gkai}她知道，\end{CJK}}  & \MYFONTSIZE{\begin{CJK}{UTF8}{gkai}这是，她知道，\end{CJK}} \\
\MYFONTSIZE{\begin{CJK}{UTF8}{gkai}这就是万物的方式。\end{CJK}}  & \MYFONTSIZE{\begin{CJK}{UTF8}{gkai}万物的运行方式。\end{CJK}} \\
\MYFONTSIZE{\begin{CJK}{UTF8}{gkai}在岸边，竹子\end{CJK}}  & \MYFONTSIZE{\begin{CJK}{UTF8}{gkai}岸上，竹子们\end{CJK}} \\
\MYFONTSIZE{\begin{CJK}{UTF8}{gkai}拍手欢呼\end{CJK}}  & \MYFONTSIZE{\begin{CJK}{UTF8}{gkai}为这般优雅\end{CJK}} \\
\MYFONTSIZE{\begin{CJK}{UTF8}{gkai}这样的优雅\end{CJK}}  & \MYFONTSIZE{\begin{CJK}{UTF8}{gkai}鼓掌，\end{CJK}} \\
\MYFONTSIZE{\begin{CJK}{UTF8}{gkai}随着风摇摆的双手\end{CJK}}  & \MYFONTSIZE{\begin{CJK}{UTF8}{gkai}随风摇摆的手\end{CJK}} \\
\MYFONTSIZE{\begin{CJK}{UTF8}{gkai}鼓掌。\end{CJK}}  & \MYFONTSIZE{\begin{CJK}{UTF8}{gkai}在风中相互拍击。\end{CJK}} \\
\hline
\MYFONTSIZE{\textbf{EAPMT-3.5}}  & \MYFONTSIZE{\textbf{EAPMT-4}} \\
\MYFONTSIZE{\begin{CJK}{UTF8}{gkai}平衡\end{CJK}}  & \MYFONTSIZE{\begin{CJK}{UTF8}{gkai}平衡\end{CJK}} \\
\PARASPACE{ } &\PARASPACE{ } \\
\MYFONTSIZE{\begin{CJK}{UTF8}{gkai}白鹤\end{CJK}}  & \MYFONTSIZE{\begin{CJK}{UTF8}{gkai}白鹤\end{CJK}} \\
\MYFONTSIZE{\begin{CJK}{UTF8}{gkai}用一条难以置信的腿\end{CJK}}  & \MYFONTSIZE{\begin{CJK}{UTF8}{gkai}以一脚不可思议地\end{CJK}} \\
\MYFONTSIZE{\begin{CJK}{UTF8}{gkai}根植于她的水镜中。\end{CJK}}  & \MYFONTSIZE{\begin{CJK}{UTF8}{gkai}稳稳扎根于她的液态镜面。\end{CJK}} \\
\MYFONTSIZE{\begin{CJK}{UTF8}{gkai}她感受着水\end{CJK}}  & \MYFONTSIZE{\begin{CJK}{UTF8}{gkai}她感觉到水在它注定的\end{CJK}} \\
\MYFONTSIZE{\begin{CJK}{UTF8}{gkai}在它注定的\end{CJK}}  & \MYFONTSIZE{\begin{CJK}{UTF8}{gkai}通往大海的旅程中\end{CJK}} \\
\MYFONTSIZE{\begin{CJK}{UTF8}{gkai}流向大海的旅途中\end{CJK}}  & \MYFONTSIZE{\begin{CJK}{UTF8}{gkai}流过她波动的倒影。\end{CJK}} \\
\MYFONTSIZE{\begin{CJK}{UTF8}{gkai}在自己泛起的波纹下穿行。\end{CJK}}  & \MYFONTSIZE{\begin{CJK}{UTF8}{gkai}这，她知道，\end{CJK}} \\
\MYFONTSIZE{\begin{CJK}{UTF8}{gkai}她明白，\end{CJK}}  & \MYFONTSIZE{\begin{CJK}{UTF8}{gkai}是万物的运行之道。\end{CJK}} \\
\MYFONTSIZE{\begin{CJK}{UTF8}{gkai}这就是事物的运行方式。\end{CJK}}  & \MYFONTSIZE{\begin{CJK}{UTF8}{gkai}岸上，竹子\end{CJK}} \\
\MYFONTSIZE{\begin{CJK}{UTF8}{gkai}在岸边，竹子\end{CJK}}  & \MYFONTSIZE{\begin{CJK}{UTF8}{gkai}以摇曳的手掌\end{CJK}} \\
\MYFONTSIZE{\begin{CJK}{UTF8}{gkai}用巨大的鼓掌\end{CJK}}  & \MYFONTSIZE{\begin{CJK}{UTF8}{gkai}为这般优雅\end{CJK}} \\
\MYFONTSIZE{\begin{CJK}{UTF8}{gkai}向如此的优雅\end{CJK}}  & \MYFONTSIZE{\begin{CJK}{UTF8}{gkai}鼓掌\end{CJK}} \\
\MYFONTSIZE{\begin{CJK}{UTF8}{gkai}鼓掌致敬，\end{CJK}}  & \MYFONTSIZE{\begin{CJK}{UTF8}{gkai}在风中拍打。\end{CJK}} \\
\MYFONTSIZE{\begin{CJK}{UTF8}{gkai}随风摇曳的双手。\end{CJK}}  &  \\
\hline
\end{tabular}
\renewcommand{\arraystretch}{1.0}
\caption{\label{Examples of Poem Translation}
This table showcases an instance of the poem "Balance," featuring the original English text, its target Chinese translation, translations generated using optimal prompts identified by the models, and the results produced by our proposed Explanation-Assisted Poetry Machine Translation (EAPMT) method.
}
\end{table*}




\subsection{Examples of the EAPMT method}
\label{Examples of the EAPMT method}


Taking the English poem ``Balance'' as an example, the prompts, explanations, and results of the Explanation-Assisted Poetry Machine Translation (EAPMT) through GPT-4 are illustrated in Table \ref{Prompt for generating explanation in Step 1}, Table \ref{Explanation generated by GPT-4}, Table \ref{Prompt for translation in Step 2}, and Table \ref{Chinese translation generated by GPT-4} respectively.

\begin{table*}
\centering
\renewcommand{\arraystretch}{0.9}
\begin{tabular}{p{15cm}}
\hline
\vspace{-0.005cm}
\MYFONTSIZE{{Please provide an explanation for this poem:}}  \\
\vspace{-0.005cm}
\MYFONTSIZE{{Balance}}  \\
\vspace{-0.005cm}
\MYFONTSIZE{{The white crane}}  \\
\MYFONTSIZE{{anchors to her liquid mirror}}  \\
\MYFONTSIZE{{on one improbable leg.}}  \\
\MYFONTSIZE{{She feels water on its ordained}}  \\
\MYFONTSIZE{{journey to the sea}}  \\
\MYFONTSIZE{{pass beneath her rippling image.}}  \\
\MYFONTSIZE{{It is, she knows,}}  \\
\MYFONTSIZE{{the way of things.}}  \\
\MYFONTSIZE{{On the shore, bamboo}}  \\
\MYFONTSIZE{{applauds}}  \\
\MYFONTSIZE{{such grace}}  \\
\MYFONTSIZE{{with swaying hands}}  \\
\MYFONTSIZE{{clapping in the wind.}}  \\
\vspace{-0.005cm}
\MYFONTSIZE{{Explanation:}}  \\
\hline
\end{tabular}
\renewcommand{\arraystretch}{1.0}
\caption{\label{Prompt for generating explanation in Step 1}
Step (1) Prompt for Generating Poem Explanations: This table presents the prompts used to guide models in generating explanations for poems, laying the foundation for subsequent translation tasks.}
\end{table*}

\begin{table*}
\centering
\renewcommand{\arraystretch}{0.9}
\begin{tabular}{p{15cm}}
\hline
\vspace{-0.005cm}
\MYFONTSIZE{{``Balance'' is a poem that draws upon the natural world to convey themes of stability, harmony, and the acceptance of life's flow. Here's a breakdown of the poem's imagery and meaning:}}  \\
\vspace{-0.005cm}
\MYFONTSIZE{{``The white crane}}  \\
\MYFONTSIZE{{anchors to her liquid mirror}}  \\
\MYFONTSIZE{{on one improbable leg.''}}  \\
\vspace{-0.005cm}
\MYFONTSIZE{{The opening lines introduce a white crane standing on one leg, a common position for cranes, which they maintain for various reasons, including conserving body heat and resting. The crane's reflection on the water's surface is described as a ``liquid mirror,'' emphasizing the stillness and clarity of the moment. The word ``improbable'' suggests a sense of wonder or marvel at the bird's ability to maintain such a poised stance.}}  \\
\vspace{-0.005cm}
\MYFONTSIZE{{``She feels water on its ordained}}  \\
\MYFONTSIZE{{journey to the sea}}  \\
\MYFONTSIZE{{pass beneath her rippling image.''}}  \\
\vspace{-0.005cm}
\MYFONTSIZE{{The crane is acutely aware of the water moving around and under her, on its ``ordained journey to the sea.'' This line introduces the concept of destiny or natural order—the water is following a path it is meant to take. The crane's ``rippling image'' suggests that while she is a part of this environment, she is also distinct from it, her presence momentarily altering the water's surface.}}  \\
\vspace{-0.005cm}
\MYFONTSIZE{{``It is, she knows,}}  \\
\MYFONTSIZE{{the way of things.''}}  \\
\vspace{-0.005cm}
\MYFONTSIZE{{These lines affirm the crane's understanding of the natural process, accepting the inevitability and the flow of life. This acceptance is reflective of a broader philosophical or spiritual understanding of life's transient nature.}}  \\
\vspace{-0.005cm}
\MYFONTSIZE{{``On the shore, bamboo}}  \\
\MYFONTSIZE{{applauds}}  \\
\MYFONTSIZE{{such grace}}  \\
\MYFONTSIZE{{with swaying hands}}  \\
\MYFONTSIZE{{clapping in the wind.''}}  \\
\vspace{-0.005cm}
\MYFONTSIZE{{The final lines shift focus from the crane to the bamboo on the shore, personifying it as an audience that ``applauds'' the crane's grace. The bamboo's ``swaying hands'' clapping in the wind create an auditory image that complements the visual imagery of the poem. Bamboo is often associated with resilience, flexibility, and strength in Asian cultures, further enhancing the themes of balance and harmony.}}  \\
\vspace{-0.005cm}
\MYFONTSIZE{{Overall, the poem ``Balance'' uses the imagery of a crane balancing on one leg and the responsive natural environment to explore themes of equilibrium and acceptance. The crane's stillness amidst the flowing water mirrors the idea of finding stability within change, and the bamboo's applause symbolizes a recognition of the beauty and grace in living in harmony with the natural world. The poem suggests that there is an intrinsic order and beauty to life's processes, and there is grace in understanding and embodying this balance.}}  \\
\hline
\end{tabular}
\renewcommand{\arraystretch}{1.0}
\caption{\label{Explanation generated by GPT-4}
GPT-4 Generated Explanations: This table displays the explanations generated by GPT-4, illustrating the model's interpretative output for a poem.}
\end{table*}

\begin{table*}
\centering
\renewcommand{\arraystretch}{0.9}
\begin{tabular}{p{15cm}}
\hline
\vspace{-0.005cm}
\MYFONTSIZE{{Please provide the Chinese translation for this
poem based on its explanation:}}  \\
\vspace{-0.005cm}
\MYFONTSIZE{{Explanation: \{Explanation generated by
GPT-4 (Same as step 1)\}}}  \\
\MYFONTSIZE{{Poem: \{English Poem “Balance” (Same as
step 1)\}
}}  \\
\vspace{-0.005cm}
\MYFONTSIZE{{Chinese translation:}}  \\
\hline
\end{tabular}
\renewcommand{\arraystretch}{1.0}
\caption{\label{Prompt for translation in Step 2}
Step (2) Integrating Explanations into the Translation Prompt: This prompt is specifically designed to incorporate the explanations as additional information, guiding the model in accurately translating the poem by providing deeper contextual insights.
}
\end{table*}

\begin{table*}
\centering
\renewcommand{\arraystretch}{0.9}
\begin{tabular}{p{15cm}}
\hline
\vspace{-0.005cm}
\MYFONTSIZE{{\begin{CJK}{UTF8}{gkai}平衡\end{CJK}}}  \\
\vspace{-0.005cm}
\MYFONTSIZE{{\begin{CJK}{UTF8}{gkai}白鹤\end{CJK}}}  \\
\MYFONTSIZE{{\begin{CJK}{UTF8}{gkai}以一脚不可思议地\end{CJK}}}  \\
\MYFONTSIZE{{\begin{CJK}{UTF8}{gkai}稳稳扎根于她的液态镜面。\end{CJK}}} \\
\MYFONTSIZE{{\begin{CJK}{UTF8}{gkai}她感觉到水在它注定的\end{CJK}}}  \\
\MYFONTSIZE{{\begin{CJK}{UTF8}{gkai}通往大海的旅程中\end{CJK}}}  \\
\MYFONTSIZE{{\begin{CJK}{UTF8}{gkai}流过她波动的倒影。\end{CJK}}}  \\
\MYFONTSIZE{{\begin{CJK}{UTF8}{gkai}这，她知道，\end{CJK}}}  \\
\MYFONTSIZE{{\begin{CJK}{UTF8}{gkai}是万物的运行之道。\end{CJK}}}  \\
\MYFONTSIZE{{\begin{CJK}{UTF8}{gkai}岸上，竹子\end{CJK}}}  \\
\MYFONTSIZE{{ \begin{CJK}{UTF8}{gkai}以摇曳的手掌\end{CJK}}}  \\
\MYFONTSIZE{{\begin{CJK}{UTF8}{gkai}为这般优雅\end{CJK}}}  \\
\MYFONTSIZE{{\begin{CJK}{UTF8}{gkai}鼓掌\end{CJK}}}  \\
\MYFONTSIZE{{\begin{CJK}{UTF8}{gkai}在风中拍打。\end{CJK}}}  \\
\hline
\end{tabular}
\renewcommand{\arraystretch}{1.0}
\caption{\label{Chinese translation generated by GPT-4}
Final Translation Result Produced by GPT-4: This entry provides the final translation outcome as generated by the GPT-4 model, showcasing its ability to interpret and render the original poem into the target language.
}
\end{table*}

\subsection{GPT-4 Evaluation Details}
\label{subsec:GPT-4 evaluation details}
In addition to human evaluation, we conducted parallel evaluation experiments using GPT-4 to determine the consistency of the results with those obtained from human assessors. The experimental setup, including the specifically designed prompts, is detailed in Table \ref{GPT-4 Evaluation Prompts}. Table \ref{GPT-4 evaluation results} presents a comparative analysis of the evaluation results for our proposed method against traditional methods and an online translation system, demonstrating the superiority of our approach.

\begin{table*}
\centering
\renewcommand{\arraystretch}{0.9}
\begin{tabular}{p{15cm}}
\hline
\MYFONTSIZE{\textbf{GPT-4 Evaluation Prompt Template}}  \\ 
\vspace{-0.005cm}
\MYFONTSIZE{Please evaluate the following five candidate translations based on eight criteria: Overall Impression, Similarity, Fidelity, Line-breaking, Meaningfulness, Poeticity, Accuracy, and Errors. In this context, English poetry serves as the source language, and the reference translation is considered the gold standard. The score for each criterion should range from 1 to 6, with higher scores indicating superior translation quality. A score of 5 signifies that the translation is of comparable quality to the reference translation, while a score of 6 indicates that the translation surpasses the quality of the reference translation.}  \\
\MYFONTSIZE{{Each criterion is defined as follows: (Same as human evaluation criteria)}}  \\
\MYFONTSIZE{{Source Language Poem: \{\{$X_i$\}\}}}  \\
\MYFONTSIZE{{Reference translation: \{\{$Y_i$\}\}}}  \\
\MYFONTSIZE{{Candidate translation 1: \{\{$C_i{}_1$\}\}}}  \\
\MYFONTSIZE{{Candidate translation 2: \{\{$C_i{}_2$\}\}}}  \\
\MYFONTSIZE{{Candidate translation 3: \{\{$C_i{}_3$\}\}}}  \\
\MYFONTSIZE{{Candidate translation 4: \{\{$C_i{}_4$\}\}}}  \\
\MYFONTSIZE{{Candidate translation 5: \{\{$C_i{}_5$\}\}}}  \\
\MYFONTSIZE{{The scores of different candidate translations under various criteria: }}  \\
\hline
\end{tabular}
\renewcommand{\arraystretch}{1.0}
\caption{\label{GPT-4 Evaluation Prompts}
The prompts we used for GPT-4 Evaluation.
}
\end{table*}




    
    
   
    
    
    


\begin{table*}
\centering
\begin{tabular}{llllllllll}
\hline
\textbf{} & {OI} & {Sim} & {Fide} & {Line} & {Mean} & {Poet} & {Acc} & {Erro} \\
\hline
{Online System} & 3.10 & 3.20 & 3.30 & 4.00 & 3.20 & 3.10 & 2.70 & 3.20 & \textbf{} \\
\hdashline
{GPT3.5-Best} & 4.20 & \textbf{4.70} & 4.60 & 5.00 & 4.60 & 4.40 & \textbf{4.50} & \textbf{4.70} & \textbf{} \\
{EAPMT-3.5} & \textbf{4.30} & 4.60 & \textbf{4.60} & \textbf{5.00} & \textbf{4.60} & \textbf{4.40} & 4.30 & 4.60 & \textbf{} \\
\hdashline
{GPT4-Best} & 4.40 & \textbf{4.80} & 4.70 & 5.00 & 4.60 & 4.50 & 4.60 & \textbf{4.90} & \textbf{} \\
{EAPMT-4.0} & \textbf{4.70} & 4.70 & \textbf{4.80} & \textbf{5.00} & \textbf{4.80} & \textbf{4.80} & \textbf{4.70} & 4.80 & \textbf{} \\
\hline
\end{tabular}
\caption{\label{GPT-4 evaluation results}
Comparison of GPT-4 evaluation results for our proposed method, traditional methods, and the online translation system, Baidu Translate.
}
\end{table*}

\end{document}